\def\B{{\mathbf B}}

\def\L{{\mathbf L}}
\def\O{{\mathbf O}}

\def\F{{\mathbf F}}

\def\L{{\cal L}}

\def\H{{\mathbf H}}
\def\F{{\mathbf T}}
\def\D{{\mathbf D}}
\def\P{{\mathbf C}}
\def\Q{{\mathbf N}}
\def\K{{\mathbf G}}
\def\H{{\mathbf M}}
\def\R{{\mathbf R}}
\def\I{{\mathbf I}}

\documentclass[journal]{IEEEtran}
\usepackage{graphicx}
\usepackage{float}
\usepackage{amsmath,graphicx,mathtools}
\usepackage{amsfonts}
 \usepackage{multirow}
 \usepackage[table,xcdraw]{xcolor}
\usepackage{algorithm}
\usepackage{algpseudocode}
\begin{document}

\title{Hy-Tracker: A Novel Framework for Enhancing Efficiency and Accuracy of Object Tracking in Hyperspectral Videos}

\author{Mohammad~Aminul~Islam,~\IEEEmembership{}
        Wangzhi Xing,~\IEEEmembership{}
        Jun~Zhou,~\IEEEmembership{Senior Member, ~IEEE},
        ~Yongsheng Gao,~\IEEEmembership{Senior Member,~IEEE}, and
        Kuldip K. Paliwal
\thanks{Mohammad Aminul Islam is with the School of Information and Communication Technology, Griffith University, Australia and the Department of Computer Science and Mathematics, Bangladesh Agricultural University, Mymensingh-2202 (mohammadaminul.islam@griffithuni.edu.au)}
\thanks{Wangzhi~Xing and Jun~Zhou are with the School of Information and Communication Technology, Griffith University, Australia ( w.xing@griffith.edu.au, jun.zhou@griffith.edu.au)}
\thanks{Yongsheng Gao and Kuldip K. Paliwal are with the Institute for Integrated and Intelligent Systems, Griffith University, Australia (yongsheng.gao@griffith.edu.au, k.paliwal@griffith.edu.au )}
\thanks{Manuscript received -----; revised September ------.}}

\markboth{Journal of \LaTeX\ Class Files,~Vol.~13, No.~9, September~2014}%
{Shell \MakeLowercase{\textit{et al.}}: Bare Demo of IEEEtran.cls for Journals}

\maketitle

\begin{abstract}
Hyperspectral object tracking has recently emerged as a topic of great interest in the remote sensing community. The hyperspectral image, with its many bands, provides a rich source of material information of an object that can be effectively used for object tracking. While most hyperspectral trackers are based on detection-based techniques, no one has yet attempted to employ YOLO for detecting and tracking the object. This is due to the presence of multiple spectral bands, the scarcity of annotated hyperspectral videos,  and YOLO's performance limitation in managing occlusions, and distinguishing object in cluttered backgrounds. Therefore, in this paper, we propose a novel framework called Hy-Tracker, which aims to bridge the gap between hyperspectral data and state-of-the-art object detection methods to leverage the strengths of YOLOv7 for object tracking in hyperspectral videos. Hy-Tracker not only introduces YOLOv7 but also innovatively incorporates a refined tracking module on top of YOLOv7. The tracker refines the initial detections produced by YOLOv7, leading to improved object-tracking performance. Furthermore, we incorporate Kalman-Filter into the tracker, which addresses the challenges posed by scale variation and occlusion. The experimental results on hyperspectral benchmark datasets demonstrate the effectiveness of Hy-Tracker in accurately tracking objects across frames.

\end{abstract}


\IEEEpeerreviewmaketitle

\section{Introduction}
Object tracking is one of the fundamental and ongoing research topics in computer vision and remote sensing and has many applications in traffic monitoring~\cite{cerutti2012optimum}, autonomous vehicles~\cite{gao2023cbff}, and video surveillance~\cite{shao2019tracking}, etc. The primary objective of object tracking is to determine the size and location of a specific object and accurately trace its trajectory across consecutive video frames, relying on the initial position and size of the target object~\cite{yang2023siammdm}. Traditional object trackers designed on RGB videos encounter difficulties in some challenging scenarios, including background clutters, similar color and texture of objects and their background, the close appearance of multiple objects in the same category, and significant deformation~\cite{su2022gaussian}. In contrast, hyperspectral videos offer a significant advantage in these scenarios as they capture a broader range of light wavelengths and higher spectral resolution compared to RGB videos~\cite{wang2022spectral}. Each pixel in hyperspectral videos contains detailed spectral information, enabling precise analysis and identification of materials and objects~\cite{yu2021feedback}. This distinctive spectral signature empowers the tracker to better follow objects~\cite{zhao2022ranet}, provide higher discrimination capabilities, and enhance object tracking accuracy in challenging scenarios~\cite{zhao2022tftn}.

Hyperspectral trackers can be broadly categorized into two groups~\cite{li2023siambag}: correlation filtering (CF) based methods and deep learning-based methods. Correlation filtering-based methods rely on correlation filters to establish a relationship between the object's features and the surrounding context in the hyperspectral video. Qian et al.~\cite{qian2018object} first combined convolutional networks with the Kernelized Correlation Filter (KCF)~\cite{henriques2014high} framework to track objects in hyperspectral videos. They used normalized three-dimensional cubes as fixed convolution filters to encode local spectral-spatial features and then applied KCF for tracking. Xiong et al.~\cite{xiong2020material} extracted material information of the target object using the spectral-spatial histogram of multidimensional gradients (SSHMGs) as 3D local spectral-spatial structures and fractional abundances of material components. Discriminative spatial-spectral features can also be extracted using spatial-spectral convolution kernels in the Fourier transform domain~\cite{chen2021object}, histogram of oriented mosaic gradient descriptors~\cite{chen2022histograms}, or adaptive spatial-spectral discriminant analysis~\cite{tang2022target}. Hou et al.~\cite{hou2022spatial} constructed a CF-based framework using a tensor sparse correlation filter (CF) with a spatial-spectral weighted regularizer to reduce spectral differences in homogeneous backgrounds and to penalize filter templates based on spectral dissimilarity.

In contrast, deep learning-based methods extract discriminative features and intricate patterns directly from hyperspectral data, showcasing notable performance within hyperspectral object tracking. Uzkent et al.~\cite{uzkent2017aerial} introduced an online generative hyperspectral tracking method that relies on likelihood maps to aid tracking without the need for offline classifiers or extensive hyperparameter tuning. They further used Kernelized Correlation Filters (KCF) and Deep Convolutional Neural Network (CNN) features for aerial object tracking in the hyperspectral domain~\cite{ uzkent2018tracking}. Liu et al.~\cite{liu2021anchor} introduced the HA-Net, employing an anchor-free Siamese network that includes a spectral classification branch. This approach leverages all hyperspectral bands during training, significantly augmenting object identification capabilities. Subsequently, Liu et al.~\cite{liu2021unsupervised} presented the H3-Net, a framework designed for high spectral-spatial-temporal resolution hyperspectral tracking. This approach encompasses an unsupervised training strategy and a dual-branch Siamese network structure. The fusion of deep learning and discriminative correlation filters (DCFs) within the H3-Net framework further enhances feature compatibility and overall tracking performance.

While these deep learning-based trackers exhibit promising performance in the hyperspectral domain, their performance is limited by insufficient training datasets. To address this challenge, Li et al.~\cite{li2020bae} introduced BAE-Net through an encoder and decoder architecture, which splits the hyperspectral bands into multiple three-channel pseudo-color images based on band-wise weights generated by the band attention model. These pseudo-color images are then fed into the adversarial learning-based tracker VITAL~\cite{song2018vital}, producing several weaker trackers that are subsequently combined through ensemble learning to determine the precise object location. In another method, they~\cite{li2021spectral} split the hyperspectral bands into multiple pseudo-color images based on the score of the spatial-spectral-temporal attention module. However, it is noted that the weak trackers in~\cite{li2020bae} and~\cite{li2021spectral} partially extract the object spectral information and thus limit their ability to capture highly discriminative features. Therefore, SiamF~\cite{li2022material} is introduced by fusing material information with the Siamese network. SiamF incorporates a hyperspectral feature fusion (HFF) module characterized by a dense connection architecture. This module integrates features from various layers and band groups, employing global-local channel attention to create a comprehensive spatial-spectral representation. Additionally, SiamF incorporates online spatial and material classifiers for adaptive online tracking. Tang et al.~\cite{tang2023siamese}  introduced a heterogeneous encoder-decoder (HED) and spectral semantic representation (SSR) modules. These modules are designed for the extraction of spatial and spectral semantic features. They also adopted a two-stage training approach to learn the relevant parameters. The acquired spatial-spectral representations are subsequently merged to estimate the optimal target state.

\begin{figure*}[t]
 \centering
  \includegraphics[width=0.9\textwidth]{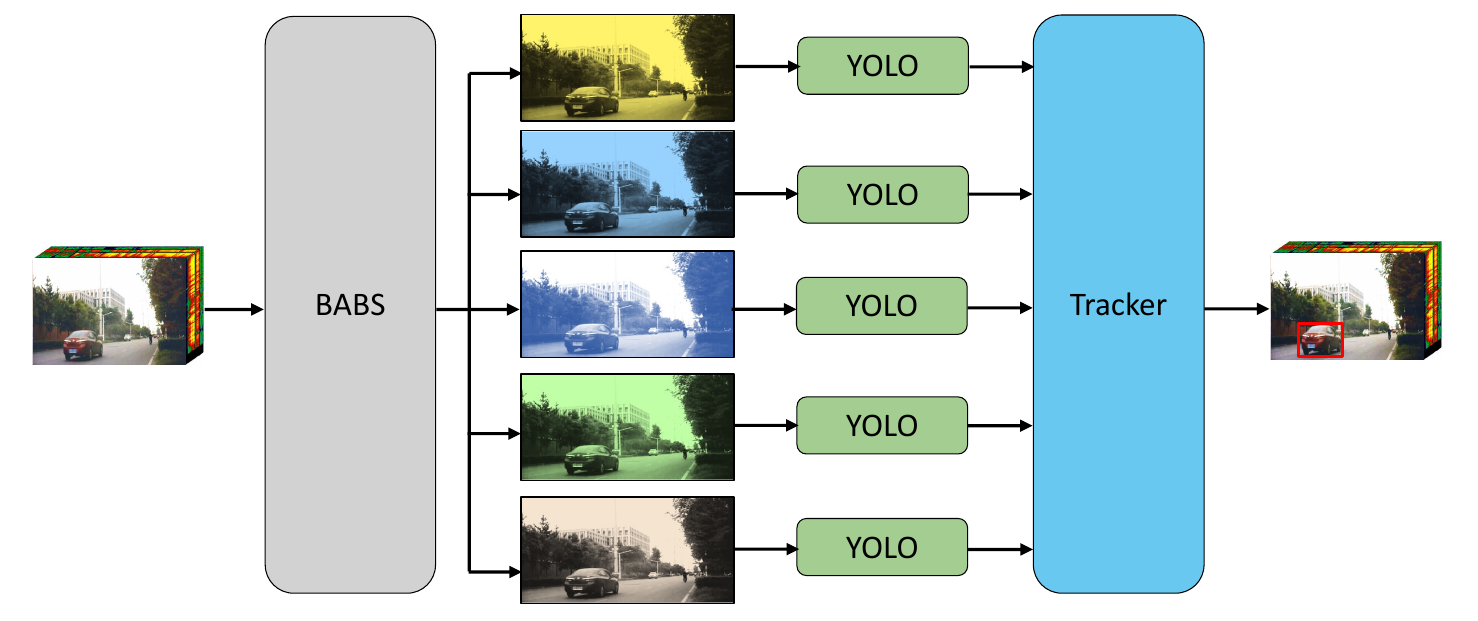}
  \caption{This figure illustrates the architecture of HyTracker, which comprises three main modules: Background Aware Band Selection (BABS), YOLO Detection, and a tracker. At first, the BABS is designed to split the bands and then regroup them into multiple three-channel pseudo-color images. After that, candidate proposals are generated by sending the images to the YOLO. Finally, these proposals undergo a refinement process using an advanced tracker, which is capable of handling challenges, including occlusions, fast motion, and background clutter.}
  \label{fig:model_architecture}
  \vspace{-10pt}
\end{figure*}

Deep learning models demonstrate the capacity to acquire hierarchical data, enabling them to autonomously extract relevant features from the datasets. To learn these intricate features, a substantial amount of data is required. Models trained on limited datasets often manifest overfitting, resulting in memorizing the training data. Furthermore, the models trained solely on hyperspectral data, neglecting the valuable information insights in RGB videos. To bridge the gap, Liu et al.~\cite{liu2022siamhyper} introduced SiamHYPER, a dual deep Siamese network framework, incorporating a pretrained RGB-based tracker and a hyperspectral target-aware module. Li et al. ~\cite{li2023siambag} developed a SiamBAG framework, where the model undergoes training using extensive RGB data instead of Hyperspectral data. They initially employ a band attention module to group hyperspectral bands into multiple pseudo-color images. These images are then processed by SiamBAG, which combines classification responses from various branches to improve object localization. Lei et al.~\cite{lei2022spatial} introduced SSDT-Net by employing transfer learning to adapt knowledge from traditional color videos for hyperspectral tracking. It incorporates a dual-transfer strategy to evaluate the similarity between the source and target domains, thereby optimizing the utilization of deep-learning models. Tang et al.~\cite{tang2022robust} designed a BAHT, which describes the semantic features of target appearances by utilizing pretrained backbone networks that had been trained on color videos.

To obtain better semantic information, some researchers used both RGB and hyperspectral data~\cite{zhao2022tftn, zhao2022ranet}. Zhao et al.~\cite{zhao2022tftn} obtained modality-specific features from RGB and hyperspectral data using a dual-branch Siamese network and transformer. This method was extended to a dual branch transformer tracking (RANet) architecture, guided by modality reliability to ensure the most informative feature is used to improve tracking accuracy. However, data from different viewpoints may not be geometrically well aligned. In certain instances, some objects or areas of interest may be visible {from one viewpoint but not from the other. This discrepancy makes it challenging to fuse information effectively. Therefore, there is a pressing need for a framework that can effectively address data dependency and increase tracking accuracy. 

YOLO~\cite{redmon2016you} is renowned for its robust object detection capabilities in RGB images. However, surprisingly, to the best of our knowledge, YOLO has not been applied for object tracking in hyperspectral videos. This omission is primarily because applying YOLO to hyperspectral data is not a straightforward task due to the multiple spectral bands, occlusions, and inherent performance constraints when adapting to hyperspectral data. Therefore, in this paper, we introduce Hy-tracker, a novel tracking method leveraging YOLO to the hyperspectral tracking domain for the very first time. Our contributions are five-fold:

\begin{enumerate}
    \item A novel tracking framework Hy-Tracker is introduced, which leverages the strengths of YOLOv7~\cite{wang2023yolov7} for the very first time and addresses the unique challenges posed by hyperspectral tracking. 
    \item A refined tracking module is designed on top of YOLO, which addresses the limitations and challenges associated with YOLO, including occlusion, background clutter, fast motion, and unseen objects. This refinement is essential for addressing these challenges and ensuring that object tracking remains accurate and reliable.
    \item We integrate the Kalman filter into the tracker which helps to handle challenges such as scale variation and occlusion, and better explore the time domain correlations between the same object in different frames. 
    \item The experimental results on hyperspectral benchmark datasets validate the robustness and effectiveness of Hy-Tracker in achieving accurate object tracking from hyperspectral videos.
\end{enumerate}

The remaining of this paper is organized as follows. We provide an in-depth discussion of the proposed tracking framework in Section II. In Section III, we present and discuss the experimental results, including the datasets, the evaluation metrics, and the performance of Hy-Tracker in various tracking scenarios. Finally, we draw conclusions in Section IV.

\section{Hy-Tracker Framework}
In this section, we will discuss the proposed Hy-Tracker framework. The Hy-Tracker framework divides a Hyperspectral Video Frame (HVF) into numerous three-channel pseudo-color images based on the band-wise cross-band dissimilarity scores. These pseudo-color images are then fed into the YOLOv7 for generating the candidate's proposal. After that, the candidates are refined to pinpoint the accurate object size and location. 

The architecture of the proposed Hy-Tracker is shown in Fig.~\ref{fig:model_architecture}. As shown in the figure, this framework consists of three main modules: Background Aware Band Selection, YOLO, and a tracker. Each of these modules is described below.

\begin{figure}[t]
 \setlength{\abovecaptionskip}{0pt}
 \setlength{\belowcaptionskip}{-15pt}
 \centering
  \includegraphics[width=0.95\columnwidth]{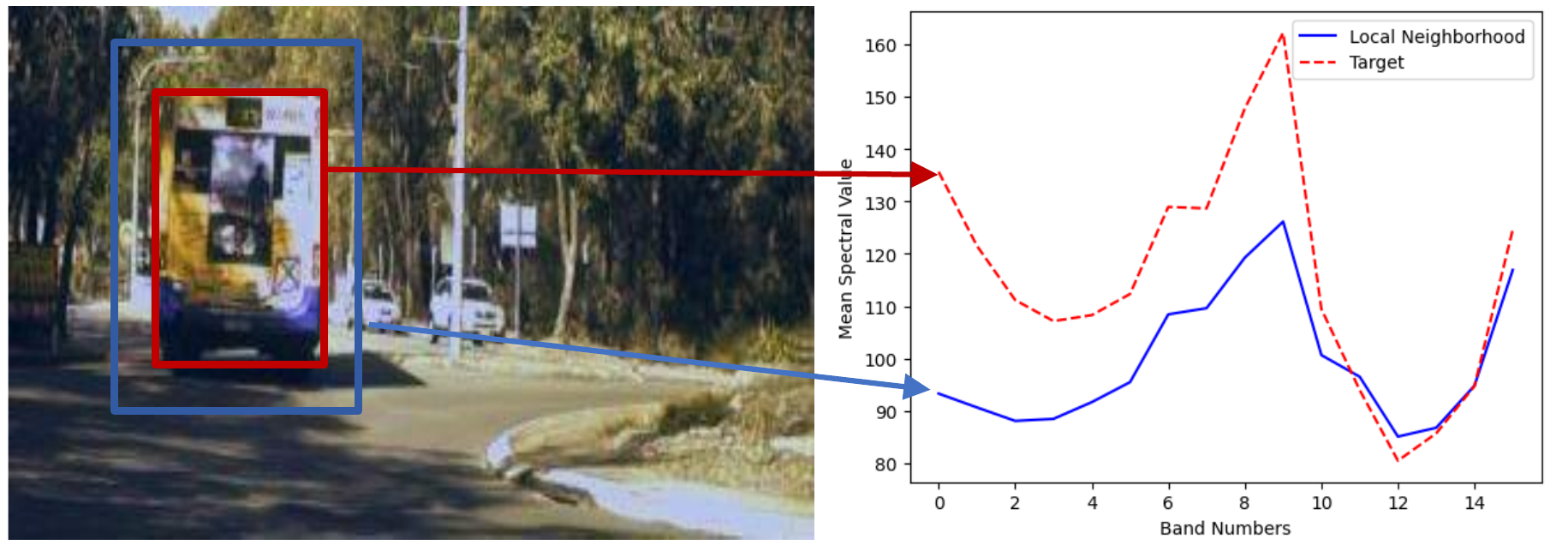}
  \caption{The average spectral contrast between the object and the adjacent local neighborhood area.}
  \label{fig:spectral_difference}
  \vspace{-10pt}
\end{figure}

\subsection{Background Aware Band Selection (BABS)}
The YOLO object detection method is renowned for its reliability but encounters challenges when directly applied to the hyperspectral domain due to the presence of an extensive number of spectral bands. Each hyperspectral band holds unique information about the object in the scene. Hyperspectral data often exhibits strong correlations between adjacent bands and not all of these bands are equally important for object tracking. In fact, the same tracker may not yield consistent results across different spectral bands. This leads to the need for a 
method capable of identifying the most significant spectral bands within hyperspectral data and effectively integrating them into the computational network.

It's worth noting that different objects possess distinct physical characteristics, resulting in variations in the spectral information related to the object itself and its local surroundings. For instance, in Fig.\ref{fig:spectral_difference}, we can observe the mean spectral disparity between an object and the adjacent local surrounding area. Certain spectral bands exhibit substantially higher levels of spectral disparity in comparison to their counterparts. Therefore, images generated with these more discriminative bands hold the potential to provide unique signatures for objects. Motivated by this observation, we employ an innovative approach~\cite{islam2023background} for selecting bands utilizing the cross-band object and local neighborhood dissimilarity scores,  as shown in Fig.~\ref{fig:babs}.

\begin{figure*}[h]
 \centering
  \includegraphics[width=1\textwidth]{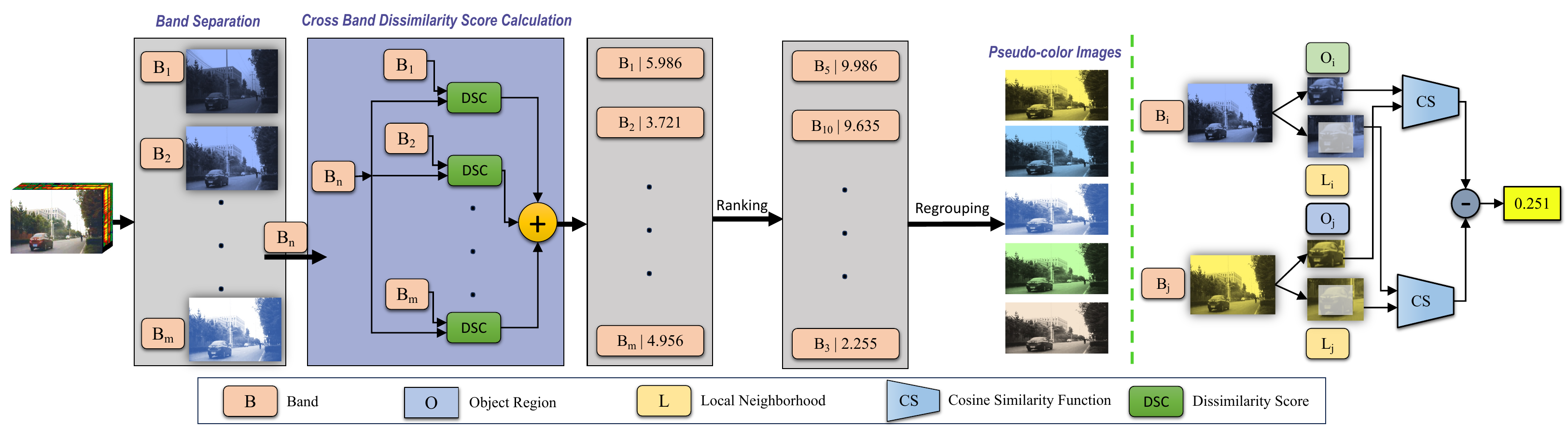}
  \caption{This figure presents the Background Aware Band Selection (BABS) module. Initially, for each band, we separate the object region and the respective local neighborhood region.  Subsequently, a dissimilarity score (DSC) is computed for every band in relation to all the other bands. This calculation is based on the differences in similarity between the object region and the local neighborhood.  These bands are subsequently ranked and regrouped to produce three-band pseudo-color images. A visual representation of DSC calculation for bands $\mathbf{B_i}$ and $\mathbf{B_j}$ is provided on the right side.}
  \label{fig:babs}
  \vspace{-10pt}
\end{figure*}

Let's consider an HVF frame denoted as $\B$ with a total of $m$ bands. These bands are labeled as $\B_1^{h\times w}, \B_2^{h\times w}, \ldots, \B_m^{h\times w}$, where $h$ and $w$ represent the height and width of the HVF frame, respectively. Two regions are extracted from each band: an object region ($\O$), and a local neighborhood region ($\L$). The region ($\O$) corresponds to the object's location as indicated by the ground truth of the object of interest. In contrast, the region ($\L$) represents the padded area around the object at a distance of $p$. Subsequently, we compute the dissimilarity score between the $\O$ and $\L$ regions of each band when compared to the $\O$ and $\L$ regions of all other bands. This computation ensures that the target object exhibits distinctiveness, resulting in a unique object signature. The following equation is used to compute the band $B_n$'s dissimilarity score:

\begin{equation}
\begin{aligned}
D_n = \sum_{j=1,\ j\neq n}^{M} ||CS(\O_n, \O_j) - CS(\L_n, \L_j)||
\end{aligned}
\label{eqn:dissimilarity}
\end{equation}

The function $CS(\cdot, \cdot)$ is a cosine similarity function used to measure the similarity between two image patches~\cite{nguyen2010cosine}.


Finally, the bands are ranked  based on their dissimilarity scores, to produce multiple pseudo-color images. For instance, if an HVF frame contains 16 bands, and if the band order after ranking is $\{7,10, 13, 15, 1,5, 4, 14, 9, 12, 2, 6, 3, 8, 11, 16\}$, then the first pseudo-color image is produced by bands $\{7, 10, 13\}$, the second image using bands $\{15, 1, 5\}$,  and so on. Subsequently, these images are passed to YOLO for candidate generations. 

\subsection{YOLO}
YOLO is a real-time object detection framework and was originally introduced by Redmon et al.~\cite{redmon2016you} in 2016. It divides an image into a grid of cells, where each cell is responsible for predicting objects that are located within its boundaries. 
In our framework, we integrate YOLOv7~\cite{wang2023yolov7} in our architecture. The subsequent section outlines the fundamental principles of YOLOv7.

\textbf{Model re-parameterization:} YOLOv7 makes a specific adjustment to mitigate performance drops caused by model re-parameterization. Model re-parameterization involves combining multiple modules into a single module during the inference phase. YOLOv7 removes the identity connection in concatenation-based models which allows the direct flow of information from one layer to another without any transformation. Removing these connections helps counteract the negative impact of model simplification on performance. 

\textbf{Coarse-to-fine label assignment:} YOLOv7 employs auxiliary heads to initially detect potential regions in the image where objects might be present. These auxiliary heads generate coarse labels and prioritize high recall, aiming to identify as many object candidates as possible. YOLOv7 adds auxiliary heads in the middle part of the network and a lead head in the final stage.

\textbf{Network Topology:} YOLOv7 changes the conv-activation-bn to conv-bn-activation. Conv-bn-activation normalizes the outputs of the convolutional layer before they are passed through the activation function, which reduces the number of operations in inference. Compared with conv-activation-bn, this topology stabilizes the activations and makes the network more robust to different input distributions by merging the convolution and batch normalization operations into a single operation. This is because only activation shares the non-linearity compared with conv and bn.

\textbf{Implicit knowledge transfer:} Pre-computing a vector in the training of YOLOR to serve as the implicit knowledge helper in the training of YOLOv7. It is added or multiplied to the feature map of YOLOv7 to leverage both the inherent understanding from YOLOR and the spatial patterns captured by the convolutional layers.

\textbf{Exponential Moving Average (EMA) model:} Teacher model serves as the guidance for the student model, and leads to more stable and robust performance, especially when dealing with noisy data or when the model's performance fluctuates during training. To do this, YOLOv7 first creates the student and teacher model while initializing the teacher's model in an EMA manner,  then trains student models on a labeled dataset and makes predictions on unlabelled data for both student and teacher models. The prediction made by the teacher model serves as the soft label to ensure consistency between the output of the two models.  Updating the student model in a back-propagation manner and using the teacher model to be the EMA of the student's weight. By encouraging the student model to align its predictions with the teacher's on unlabeled data, the student can benefit from the patterns and structures in the unlabeled data without overfitting to any potential noise or anomalies.

\subsection{Trackers}
The primary goal of YOLO is to detect all objects in a scene,  while in our work, object tracking focuses on following a specific object of interest across multiple frames in a video sequence.  The emphasis of the tracking is to maintain the identity and position of the object over time, which however is challenging in scenarios such as the similar appearance of multiple objects, background clutter, occlusions, low spatial resolution, etc. To address these challenges, a tracker is essential to bridge YOLO's detection and the requirement of object tracking. The tracker will transform the YOLO's detection into a coherent and reliable tracking system by handling the complexity of the dynamic scenes. Therefore, we introduce the tracker which comprises three main modules: a classifier, a target proposal generator, and a Kalman filter. 

\subsubsection{Classifier} The purpose of the classifier is to separate the object of interest from other objects and background. The classifier architecture is presented in TABLE~\ref{tab:classifier_description}.  It comprises of three convolutional layers (conv1-3) and three fully connected layers (fc1-3) which are responsible for learning and extracting features from input. The reason for choosing a smaller network is that we are refining and fine-tuning the output of YOLO, and it is computationally efficient and suitable for online training. 

\begin{table}[]
\caption{Classification network}
\begin{tabular}{ccccc}
\hline
Name  & Input         & Kernel Size & Stride & Output        \\ \hline
conv1 & 107 x 107 x 3 & 7 x 7       & 2      & 11 x 11 x 256 \\
conv2 & 11 x 11 x 256 & 5 x 5       & 2      & 5 x 5 x 256     \\
conv3 & 5 x 5 x 256     & 3 x 3       & 1      & 3 x 3 x 512   \\
fc1   & 3 x 3 x 512   &             &        & 512           \\
fc2   & 512           &             &        & 512           \\
fc3   & 512           &             &        & 2             \\ \hline
\end{tabular}
\label{tab:classifier_description}
\vspace{-10pt}
\end{table}

\subsubsection{Target Proposal Generator} 
\label{sec:proposal}

The primary purpose of this step is to generate a set of potential bounding boxes that would likely include the target object within the current video frame. These potential candidates bounding boxes are generated through Gaussian distribution sampling, following the approach described in MDNET~\cite{nam2016learning}. This distribution is characterized by two main parameters: mean and covariance. The mean of the distribution is derived from the previous state of the target object which estimates the potential position of this object within the current frame. The covariance matrix is diagonal and contains specific values $\left(0.09r^2, 0.09r^2, 0.25\right)$, where $r$ represents the mean of the target object's bounding box width and height. In addition, the target object's size may be changed slightly between frames. Therefore, we adjust the scale of each candidate by multiplying with $1.05sc$ where 'sc' denotes the scaling value of the candidate box obtained from the Gaussian distribution.

\begin{table}[]
\caption{Description of HOT 2022 testing dataset}
\centering
\begin{tabular}{cccc}
\hline
\textbf{Scenes} & \textbf{Frames} & \textbf{Size} & \textbf{Attributes}    \\ \hline
ball            & 625             & 471x207x16    & SV, MB, OCC            \\
basketball      & 186             & 463x256x16    & FM, MB, OCC, LR        \\
board           & 471             & 277x140x16    & IPR, OPR, BC, OCC, SV  \\
book            & 601             & 317x148x16    & IPR, OPR, DEF          \\
bus             & 131             & 261x148x16    & LR, BC, FM             \\
bus2            & 326             & 361x167x16    & IV, SV, OCC, FM        \\
campus          & 976             & 384x157x16    & IV, SV, OCC            \\
car             & 101             & 261x148x16    & SV, IPR, OPR, OCC      \\
car2            & 131             & 351x167x16    & SV, IPR, OPR           \\
car3            & 331             & 512x256x16    & SV, LR, OCC, IV        \\
card            & 930             & 363x200x16    & IPR, BC, OCC           \\
coin            & 149             & 291x120x16    & BC                     \\
coke            & 731             & 493x207x16    & BC, IPR, OPR, FM, SV   \\
drive           & 725             & 297x142x16    & BC, IPR, OPR, SV       \\
excavator       & 501             & 500x240x16    & IPR, OPR, SV, OCC, DEF \\
face            & 279             & 446x224x16    & IPR, OPR, SV, MB       \\
face2           & 1111            & 446x224x16    & IPR, OPR, SV, OCC      \\
forest          & 530             & 512x256x16    & BC, OCC                \\
forest2         & 363             & 512x256x16    & BC, OCC                \\
fruit           & 552             & 493x232x16    & BC, OCC                \\
hand            & 184             & 314x186x16    & BC, SV, DEF, OPR       \\
kangaroo        & 117             & 385x206x16    & BC, SV, DEF, OPR, MB   \\
paper           & 278             & 446x224x16    & IPR, BC                \\
pedestrain      & 306             & 351x167x16    & IV, SV                 \\
pedestrain2     & 363             & 512x256x16    & OCC, LR, DEF, IV       \\
player          & 901             & 463x256x16    & IPR, OPR, DEF, SV      \\
playground      & 800             & 463x256x16    & SV, OCC                \\
rider1          & 336             & 512x256x16    & LR, OCC, IV, SV        \\
rider2          & 210             & 512x256x16    & LR, OCC, IV, SV        \\
rubik           & 526             & 493x207x16    & DEF, IPR, OPR          \\
student         & 396             & 438x256x16    & IV, SV                 \\
toy1            & 376             & 271x135x16    & BC, OCC                \\
toy2            & 601             & 371x171x16    & BC, OCC, SV, IV, OPR   \\
truck           & 221             & 512x256x16    & OCC, IV, SV, OV        \\
worker          & 1209            & 228x121x16    & SV, LR, BC             \\ \hline
\end{tabular}
\label{tab:dataset2022}
\vspace{-10pt}
\end{table}

\subsubsection{Kalman Filter} 
The Kalman Filter is a recursive technique employed to estimate the state of a dynamic system when dealing with noisy measurements or observations. It's particularly useful to track an object and predict its future state based on noisy sensor data~\cite{khodarahmi2023review}. There are three main steps in the Kalman Filter.

\textbf{State Estimation: } In this step, we first estimate the object's initial state, including its position, velocity, and the covariance matrix which represents the initial uncertainty in the state estimate. 

\textbf{Prediction: }The prediction step involves estimating the future state based on the current state and a state transition model. Prediction of state and covariance is given in Equations (\ref{eqn:state}) and 
 (\ref{eqn:covariance}). 
\begin{equation}
    \hat{x}_{t|t-1} = \F_t\hat{x}_{t-1|t-1} + \D_tu_t
    \label{eqn:state}
\end{equation}
\begin{equation}
    \P_{t|t-1}= \F_t\P_{t-1|t-1}\F_{t}^{T} + \Q_{t}
    \label{eqn:covariance}
\end{equation}
Here, $x$ is the state of the targeted object, $u$ is the control input, $\P$ is associated with the state covariance matrix, $\F$ denotes the state transition matrix, $\D$ represents the control input matrix, and $\Q$ stands for the noise covariance matrix.

\textbf{Update: } The update step combines the prediction with the actual sensor measurements to obtain a more accurate estimate of the state. At first, the Kalman gain is obtained by using  Equation~\ref{eqn:kalman_gain}:
\begin{equation}
    \K_{t}= \P_{t|t-1}\H_{t}^{T}\left(\H_{t}\P_{t|t-1}\H_{t}^{T} + \R_t\right)^{-1}
    \label{eqn:kalman_gain}
\end{equation}
Then, the system state and the covariance matrix are updated using Equations~(\ref{eqn:update_state}) and (\ref{eqn:update_covariance})
\begin{equation}
    \hat{x}_{t} = \hat{x}_{t|t-1} + \K_t\left(z_t - \H_t\hat{x}_{t|t-1}\right)
    \label{eqn:update_state}
\end{equation}
\begin{equation}
    \P_{t}= \left(\I - \K_t\H_t)\right) \P_{t|t-1}
    \label{eqn:update_covariance}
\end{equation}
 where, '$\K$' denotes the Kalman gain, 'z' represents the observed measurement at time 't', '$\I$' corresponds to the identity matrix, '$\H$' stands for the measurement matrix, and '$\R$' symbolizes the covariance matrix of measurement noise.

\begin{table}[]
\centering
\caption{Detail results of Hy-Tracker on HOT-2022 Dataset}
\begin{tabular}{ccccc}
\hline
Scenes         & Frames         & AUC             & DP@20Pixels     & FPS            \\ \hline
ball           & 625            & 0.697           & 0.930           & 3.460          \\
basketball     & 186            & 0.705           & 1.000           & 3.251          \\
board          & 471            & 0.794           & 1.000           & 3.518          \\
book           & 601            & 0.725           & 1.000           & 3.640          \\
bus            & 131            & 0.737           & 1.000           & 3.423          \\
bus2           & 326            & 0.833           & 1.000           & 3.398          \\
campus         & 976            & 0.494           & 0.775           & 4.097          \\
car            & 101            & 0.811           & 1.000           & 3.444          \\
car2           & 131            & 0.771           & 1.000           & 3.478          \\
car3           & 331            & 0.611           & 0.906           & 3.096          \\
card           & 930            & 0.773           & 0.998           & 3.330          \\
coin           & 149            & 0.809           & 1.000           & 4.020          \\
coke           & 731            & 0.609           & 0.979           & 3.383          \\
drive          & 725            & 0.813           & 0.977           & 3.798          \\
excavator      & 501            & 0.686           & 0.980           & 3.093          \\
face           & 279            & 0.806           & 1.000           & 3.154          \\
face2          & 1111           & 0.842           & 1.000           & 3.129          \\
forest         & 530            & 0.753           & 1.000           & 3.018          \\
forest2        & 363            & 0.632           & 0.904           & 3.123          \\
fruit          & 552            & 0.578           & 0.839           & 3.105          \\
hand           & 184            & 0.800           & 0.973           & 3.319          \\
kangaroo       & 117            & 0.732           & 1.000           & 3.619          \\
paper          & 278            & 0.795           & 1.000           & 3.086          \\
pedestrain     & 306            & 0.732           & 1.000           & 3.447          \\
pedestrain2    & 363            & 0.666           & 0.895           & 3.014          \\
player         & 901            & 0.734           & 0.962           & 3.238          \\
playground     & 800            & 0.741           & 0.995           & 3.151          \\
rider1         & 336            & 0.627           & 1.000           & 3.580          \\
rider2         & 210            & 0.658           & 1.000           & 3.309          \\
rubik          & 526            & 0.739           & 1.000           & 3.155          \\
student        & 396            & 0.758           & 1.000           & 3.074          \\
toy1           & 376            & 0.723           & 1.000           & 3.632          \\
toy2           & 601            & 0.778           & 0.953           & 3.397          \\
truck          & 221            & 0.742           & 0.977           & 3.058          \\
worker         & 1209           & 0.540           & 0.999           & 4.059          \\ \hline
\textbf{Total} & \textbf{16574} & \textbf{72.1\%} & \textbf{97.3\%} & \textbf{3.374} \\ \hline
\end{tabular}
\label{tab:details_result}
\vspace{-10pt}
\end{table}

\subsection{Offline Training and Online Updating}
\subsubsection{Offline Training} In the offline training phase, YOLO is exclusively trained using the training dataset and the initial frame of the testing dataset. To mitigate overfitting, we adopt a Gaussian cut-paste method to mimic that the object is moving across the frame. The probability distribution function for a 2D Gaussian distribution with means $\left(\mu_p, \mu_q\right)$ and standard deviations $\left(\sigma_p, \sigma_q\right)$, given a specific object position $\left(p, q\right),$ is expressed as follows:
\begin{equation}
 f\left(p, q\right) = \frac{1}{2\pi\sigma_p\sigma_q} \exp\left(-\left(\frac{\left(p - \mu_p\right)^2}{2\sigma_p^2} + \frac{\left(q - \mu_q\right)^2}{2\sigma_q^2}\right)\right)
\end{equation}

\subsubsection{Online Updating}
In the online updating process, our primary focus is on refining the classifier within the context of tracking. To do this, we adopt the pre-trained weights of \cite{nam2016learning}. It's worth noting that during online updating, only the weights (fcw1  to fcw3) associated with fully connected layers (fc1-3) are updated, while the weights (cw1 to cw3) associated with convolutions layers (conv1 to conv3) remain fixed. We use Binary Cross Entropy (BCE) Loss for the classifier, with the initial learning rate set at $0.0005$ and subsequently updated to $0.001$.

In each frame, two distinct types of predictions are generated: YOLO's predictions and Kalman's prediction. Simultaneously, a set of $N$ target candidates is taken into consideration, originating from the previous state of the target, as specified in Section~\ref{sec:proposal}. Subsequently, these target samples and YOLO's predictions are then processed by the classifier, which assigns classification scores to each of them. After that, the optimal target state is selected by finding the one with the highest classification score. If there is a scale change exceeding 5\% between the optimal target state and the previous state, then the Kalman prediction is selected as the optimal target. The Kalman filter's state is updated after processing each frame, and in addition to this, we conduct periodic updates of the classifier weights (fcw1 to fcw3) after every 10 frames. These weight updates utilize the positive and negative candidates generated from the optimal outputs of the previous 10 frames, ensuring ongoing adaptability and tracking precision.

\section{Experimental results and discussion}
In this section,  we provide details about the experimental setup, evaluation metrics, ablation studies, and a comparative analysis with state-of-the-art hyperspectral trackers. 

\subsection{Experimental Settings} 
In our study, we have employed the 2022 hyperspectral object tracking (HOT) challenges dataset\footnote{https://www.hsitracking.com/}. This dataset was captured at a frame rate of 25 FPS using an XIMEA snapshot camera with an imec hyperspectral sensor. Each frame of the dataset was originally captured in a 2-D mosaic format, with wavelengths ranging from 470nm to 620nm. The dataset consists of a total of 40 training videos and 35 testing videos. It includes both RGB and HSI-False videos of the same scenes where the HSI-False are the color videos converted from hyperspectral videos. Notably, the testing videos incorporate 11 challenging factors, which are deformation (DF), motion blur (MB), scale variation (SV), out-of-view (OV), out-of-plane rotation (OPR), occlusion (OC), background clutter (BC), in-plane rotation (IPR), fast motion (FM), illumination variation (IV), and low resolution (LR). 
The detailed description of the dataset is provided in TABLE~\ref{tab:dataset2022}. 

All experiments were carried out on a machine featuring an Intel(R) Core i7-12700 CPU, 32 GB of RAM, and an NVIDIA RTX A4000 GPU equipped with 16 GB of dedicated graphics memory.

\begin{table*}[]
\centering
\caption{Abalation Study}
\begin{tabular}{ccccccccc}
\hline
{}  & YOLO      & Tracker   & Kalman Filter & AUC  &$\Delta$AUC& {DP@20pixels} &  $\Delta${DP@20pixels}\\ \hline
{} & $\sqrt{}$  & {}        & {}            & {0.576}  & {-}            & {0.806}&{-}\\
{} & $\sqrt{}$ & {} & $\sqrt{}$            & {0.609}  & {0.0033}            & {0.868}&{0.062}\\
RGB & $\sqrt{}$ & $\sqrt{}$ & {}            & {0.656}  & {0.080}            & {0.908}&{0.102}\\

{} & $\sqrt{}$  & $\sqrt{}$ & $\sqrt{}$     & {0.666}  & {0.090}            & {0.943}&{0.137}\\ \hline
{} & $\sqrt{}$  & {}        & {}            & {0.536}  & {-}            & {0.746} &{} \\
{} & $\sqrt{}$  & {}        & $\sqrt{}$     & {0.563}  & {0.027}            & {0.797} &{0.051} \\
HSI-False & $\sqrt{}$ & $\sqrt{}$ & {}            & {0.619}  & {0.083} &{0.861}&{0.115}  \\
{} & $\sqrt{}$  & $\sqrt{}$ & $\sqrt{}$     & {0.624}  & {0.088}            & {0.864} & {0.118} & {}\\ \hline
{} & $\sqrt{}$  & {}        & {}            & {0.637}  & {-}            & {0.851} &{-}\\
{} & $\sqrt{}$  & {}        & $\sqrt{}$     & {0.646}  & {0.009}            & {0.870} &{0.019}\\
HSI & $\sqrt{}$ & $\sqrt{}$ & {}            & {0.707}  & {0.070}            & {0.952} & {0.101} \\
{} & $\sqrt{}$  & $\sqrt{}$ & $\sqrt{}$     & {0.721}  & {0.084}& {0.973} & {0.122} \\ \hline
\end{tabular}
\label{tab:abalation_study}
\vspace{-10pt}
\end{table*}

\begin{figure*}
\centering
    \includegraphics[width=.8\linewidth]{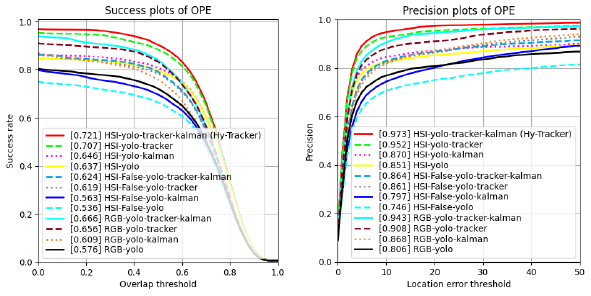}
    \vspace{-10pt}
    \caption{Ablation analysis of Hy-Tracker components in terms of success and precision plots.}
    \label{fig:ablation_study}
    \vspace{-10pt}
\end{figure*}

\subsection{Evaluation Metrices} 
In our experiments, four evaluation metrics are used, including success plots, precision plots, the area under the curve (AUC), and the precision rate at a given 20-pixel threshold (DP@20pixels). It's worth noting that the one-pass evaluation (OPE)~\cite{xiong2020material} technique was used to record all of the experimental results. 

The success plot, where the x-axis is the overlap threshold and the y-axis is the success rate (SR), shows how well a tracker performs. The success rate is determined by calculating the intersection over union (IoU) between the actual ground truth and the predicted bounding box. This measure is expressed as follows:

\begin{equation}
    SR(\tau)= \frac{N_{IoU} > \tau}{N_{total}}
\end{equation}
Here, $\tau$ represents a threshold value that ranges from 0 to 1, while 'N' denotes the number of frames.  If we denote $A_p$ as the area of the predicted representation of the target object and $A_g$ as the area of the ground truth representation, the IoU is defined as:

\begin{equation}
    IoU = \frac{A_g \cap A_p}{A_g \cup A_p}
\end{equation}
Here $\cap$ and $\cup$ indicate the IoU between the target object's ground truth and the predicted bounding boxes. A frame is considered to have been tracked successfully when the IoU exceeds the threshold $\tau$. The proximity of the curve to the upper right corner of the plot signifies superior tracker performance.

The precision plot, where the x-axis represents the center position error ($CLE$) and the y-axis corresponds to the precision rate ($PR$), provides a visual representation of a tracker's precision. The $CLE$ is calculated as the Euclidean distance between the centers of the actual and predicted bounding boxes and is defined as follows:
\begin{equation}
    CLE = \sqrt{\left(p_1 - p_2\right)^2 + \left(q_1 - q_2\right)^2}
\end{equation}
where $\left(p_1, \; q_1\right)$ and $\left(p_2,\; q_2\right)$ represent the center locations of the ground truth and predicted bounding boxes of the target object. The $CLE$ difference between the ground truth and the predicted bounding box is measured as $PR$,  which is defined as 
\begin{equation}
    PR(\tau)= \frac{N_{CLE} > \eta}{N_{total}}
\end{equation}
where $\eta$ is a threshold ranging from $0$ to $50$ pixels and $N$ represents the number of frames. A frame is considered to have been tracked successfully if the $CLE$ is less than $\eta$. The proximity of the curve to the upper left corner of the plot signifies superior tracker performance.

\begin{figure*}
\centering
    \includegraphics[width=.8\linewidth]{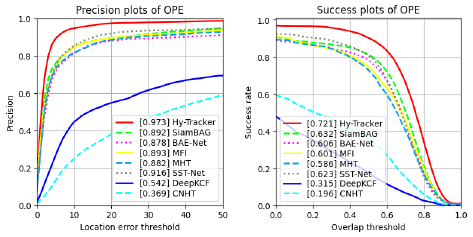}
    \vspace{-10pt}
    \caption{The comparison between different hyperspectral trackers in terms of success plot and precision plot on the HOT-2022 dataset.}.
    \label{fig:success_precision_2022}
    \vspace{-10pt}
\end{figure*}

\begin{figure*}[t]
 \setlength{\abovecaptionskip}{0pt}
 \setlength{\belowcaptionskip}{-15pt}
 \centering
  \includegraphics[width=0.9\linewidth]{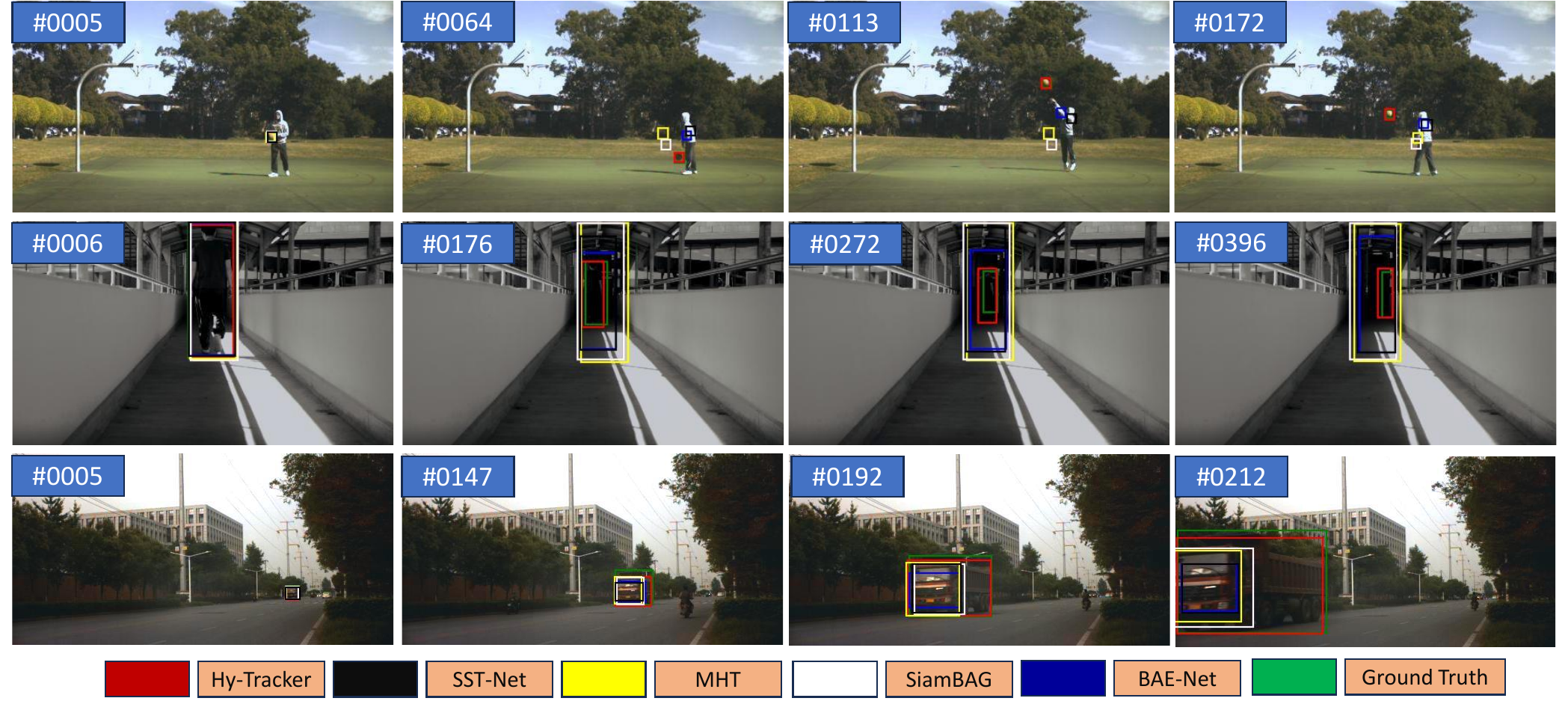}
  \caption{Demonstration of some tracking results from basketball, student, and truck videos from the HOT-2022 dataset. }
  \label{fig:demonstration2022}
  \vspace{-10pt}
\end{figure*}

\begin{figure*}
\centering
    \includegraphics[width=.8\linewidth]{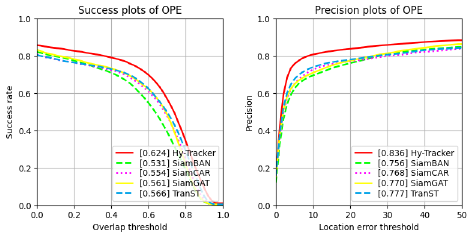}
    \vspace{-10pt}
\caption{The comparison between different hyperspectral trackers in terms of success plot and precision plot on the HOT-2023 dataset.}
\vspace{-10pt}
\label{fig:sota2023_comparison}
    \vspace{-1pt}
\end{figure*}

\begin{figure*}[t]
 \centering
  \includegraphics[width=0.9\linewidth]{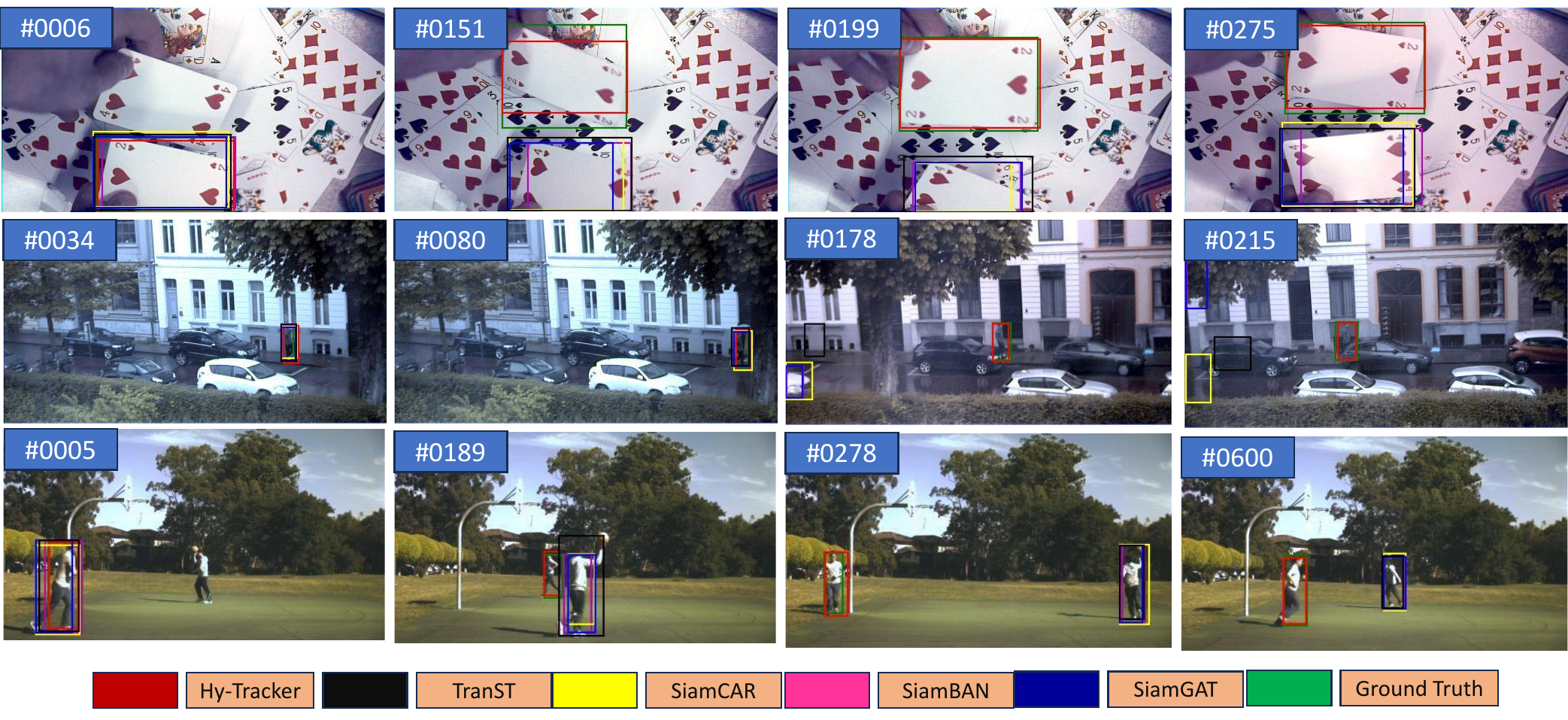}
  \caption{Demonstration of some tracking results from vis-card19, vis-rainystreet16, and vis-playground videos from the HOT-2023 dataset. }
  \label{fig:demonstration2023}
  \vspace{-10pt}
\end{figure*}

\subsection{Ablation Study} In this section, we emphasize the significant contributions of YOLO, Tracker, and Kalman filters within the Hy-Tracker architecture on all three video formats (RGB, HSI-False, and HSI). The result of this analysis is presented in TABLE~\ref{tab:abalation_study}, and the success and precision plots are depicted in Fig.~\ref{fig:ablation_study}. It is worth mentioning that we choose the highest confidence score among all the candidate proposals when we consider only YOLO for tracking. 

Referring to the results in TABLE~\ref{tab:abalation_study}, it is evident that the simultaneous integration of YOLO, Tracker, and Kalman filters allows Hy-Tracker to make substantial improvements across all video formats. In RGB videos, the AUC is improved from 0.576 to 0.666, showcasing the network's ability to excel in color-based tracking scenarios. In the case of HSI-False videos, where data is inherently noisy, Hy-Tracker still demonstrates its adaptability by increasing the AUC from $0.536$ to $0.624$. However, the most notable improvement is observed in HSI videos, wherein each object possesses a distinctive signature, where the AUC is improved from $0.637$ to $0.721$. This demonstrates that the proposed Hy-Tracker is robust and capable of adapting to a wide range of tracking conditions and video formats. The detailed results of Hy-Tracker on the HOT-2022 Dataset are presented in TABLE~\ref{tab:details_result}.

\subsection{Comparison with state-of-the-arts hyperspectral trackers} 
In this section, we conduct a comprehensive comparative analysis of Hy-Tracker in relation to several state-of-the-art tracking methods, including MHT~\cite{xiong2020material}, BAE-Net~\cite{li2020bae}, CNHT~\cite{qian2018object}, DeepKCF~\cite{uzkent2017aerial}, SST-Net~\cite{li2021spectral}, MFI~\cite{li2022material}, and SiamBAG~\cite{li2023siambag}. The detailed results of this comparative study are presented in TABLE~\ref{tab:sota2022_comparison}, and the visual representation in terms of the success and precision plots are depicted in Fig.~\ref{fig:success_precision_2022}. From these results, we observe that the Hy-Tracker achieves the highest AUC and highest precision at 20 pixels (DP@20pixels) compared to all the other trackers. The underlying reason is that MHT and CNHT rely solely on hand-crafted features, BAE-Net and SST-Net leverage hyperspectral information effectively by grouping bands into pseudo-color images and then producing several weak trackers. SiamBAG, on the other hand, benefits from extensive color object tracking data for training, resulting AUC of $0.632$, but ignores the hyperspectral dataset. In contrast, Hy-Tracker adopts a robust YOLO detection method where YOLO's detection capabilities are fine-tuned through the integration of a tracker and Kalman filter, achieving a significant \textbf{$0.721$} AUC. This integrated approach optimizes object tracking in hyperspectral scenarios, making it the most reliable and robust tracker. 

Furthermore, we conduct a detailed comparison of Hy-Tracker based on various attributes of the dataset, and the results of this comparative analysis are outlined in TABLE~\ref{tab:attribute_analysis_2022}. In each of the challenging scenarios, we observe that Hy-Tracker consistently outperforms all other trackers, demonstrating its superiority in tracking accuracy and adaptability. This demonstrates the Hy-Tracker's capacity to excel in real-world scenarios where tracking conditions can vary widely. Some of the tracking results from basketball, student, and truck are shown in Fig.~\ref{fig:demonstration2022}.

\subsection{Evalution on HOT-2023 Dataset} 
We have also conducted a comprehensive evaluation of Hy-Tracker on the latest HOT-2023 dataset. The dataset consists of three different modules of data: VIR, RedNIR, and NIR,  which contain 16, 15, and 25 bands respectively. This dataset includes 110 training videos and 87 validation videos. We compare the proposed method with four other state-of-the-art methods, including SiamBAN~\cite{chen2020siamese}, , SiamGAT~\cite{guo2021graph}, SiamCAR~\cite{guo2020siamcar} and TranST~\cite{yan2021learning}. The comparison results are shown in Fig.~\ref{fig:sota2023_comparison}. The results are taken from HOT\footnote{https://www.hsitracking.com/}. From this result, it is evident that Hy-Tracker outperforms all the state-of-the-art methods. Specifically, in terms of AUC, Hy-Tracker surpasses the second-best method, TranST, by a notable margin of 5.8\%, achieving an AUC of \textbf{0.624} compared to TranST's \textbf{0.566}. Furthermore, in terms of DP@20pixels, Hy-Tracker maintains its dominance by outperforming TranST by 5.9\%, achieving a DP@20pixels score of \textbf{0.836} compared to TranST's \textbf{0.777}. The consistent performance of Hy-Tracker across both AUC and DP@20pixels metrics demonstrates its adaptability and robustness in diverse challenging scenarios. Fig.~\ref{fig:demonstration2023} depicts some of the tracking results from the HOT-2023 dataset's vis-card19, vis-rainystreet16, and vis-playground videos, highlighting the effectiveness of Hy-Tracker in diverse real-world scenarios.

\begin{table}[]
\centering
\caption{Comparison of Hy-Tracker with different state-of-the-arts trackers. The top two results are marked by \textcolor{red}{\textbf{Red}} and \textcolor{green}{\textbf{Green}}.}
\begin{tabular}{cccc}
\hline
\multicolumn{1}{l}{} & \multicolumn{1}{l}{AUC}               & \multicolumn{1}{l}{DP@20Pixels}       & \multicolumn{1}{l}{FPS}              \\ \hline
Hy-Tracker           & {\color[HTML]{FF0000} \textbf{0.721}} & {\color[HTML]{FF0000} \textbf{0.973}} & {\color[HTML]{00B050} \textbf{3.37}} \\
SiamBAG~\cite{li2023siambag}              & {\color[HTML]{00B050} \textbf{0.632}} & 0.892                                 & {\color[HTML]{FF0000} \textbf{5.7}}  \\
BAE-Net ~\cite{li2020bae}             & 0.606                                 & 0.878                                 & \textless{}1                         \\
MFI~\cite{li2022material}                 & 0.601                                 & 0.893                                 & 2                                    \\
MHT~\cite{xiong2020material}                  & 0.586                                 & 0.882                                 & 2                                    \\
SST-Net~\cite{li2021spectral}              & 0.623                                 & {\color[HTML]{00B050} \textbf{0.916}} & \textless{}1                         \\
DeepKCF~\cite{uzkent2017aerial}              & 0.315                                 & 0.542                                 & 2.1                                  \\
CNHT~\cite{qian2018object}                 & 0.196                                 & 0.369                                 & \textless{}1                         \\ \hline
\end{tabular}
\label{tab:sota2022_comparison}
\vspace{-10pt}
\end{table}

\begin{table}[]
\centering
\caption{Comparison of different hyperspectral trackers in terms of attribute analysis of different challenging scenarios in the HOT-2022 dataset. The top results are marked by \textcolor{red}{\textbf{Red}} and \textcolor{green}{\textbf{Green}}. }
\begin{tabular}{lllllll}
\hline
    & SiamBAG & BAE-Net & MFI   & MHT   & SST-Net & Hy-Tracker \\ \hline
BC  & 0.648   & 0.651   & 0.651 & 0.606 & \textcolor{green}{\textbf{0.685}}   & \textcolor{red}{\textbf{0.724}}      \\
DEF & 0.692   & 0.680   & 0.639 & 0.664 & \textcolor{green}{\textbf{0.700}}   & \textcolor{red}{\textbf{0.726}}      \\
FM  & \textcolor{green}{\textbf{0.615}}   & 0.608   & 0.600 & 0.542 & 0.561   & \textcolor{red}{\textbf{0.721}}      \\
IPR & \textcolor{green}{\textbf{0.703}}   & 0.699   & 0.692 & 0.670 & 0.696   & \textcolor{red}{\textbf{0.761}}      \\
IV  & \textcolor{green}{\textbf{0.533}}   & 0.524   & 0.515 & 0.477 & 0.501   & \textcolor{red}{\textbf{0.690}}      \\
LR  & \textcolor{green}{\textbf{0.583}}   & 0.489   & 0.514 & 0.476 & 0.462   & \textcolor{red}{\textbf{0.649}}      \\
MB  & \textcolor{green}{\textbf{0.650}}   & 0.593   & 0.570 & 0.560 & 0.536   & \textcolor{red}{\textbf{0.735}}      \\
OCC & \textcolor{green}{\textbf{0.596}}   & 0.554   & 0.546 & 0.564 & 0.594   & \textcolor{red}{\textbf{0.707}}      \\
OPR & 0.683   & \textcolor{green}{\textbf{0.701}}   & 0.680 & 0.644 & 0.699   & \textcolor{red}{\textbf{0.760}}       \\
OV  & \textcolor{green}{\textbf{0.622}}   & 0.509   & 0.607 & 0.621 & 0.473   & \textcolor{red}{\textbf{0.742}}      \\
SV  & \textcolor{green}{\textbf{0.622}}   & 0.604   & 0.599 & 0.574 & 0.602   & \textcolor{red}{\textbf{0.722}}      \\ \hline
\end{tabular}
\label{tab:attribute_analysis_2022}
\vspace{-10pt}
\end{table}

\section{Conclusion}

In this research, a novel hyperspectral tracker, Hy-Tracker, is introduced, which bridges the gap between state-of-the-art object detection methods and hyperspectral tracking. The tracker module in the Hy-Tracker refines the YOLO's proposal and addresses the challenges of dynamic scenes. The integrated Kalman filter helps the tracker overcome the challenges of occlusion and scale variation. The promising result of the Hy-Tracker in both the HOT-2022 and HOT-2023
datasets imply the robustness and reliability of the proposed framework. However, the Kalman filter performs well when the object projection is linear. Therefore, in the future, it will be interesting to incorporate a motion-based system that will handle the non-linear projection of the object and also improve tracking accuracy and speed.

\bibliographystyle{IEEEtran}
\bibliography{bibliography} 
\end{document}